\documentclass{article}
\usepackage{microtype}
\usepackage{graphicx}
\usepackage{tikz}
\usepackage{enumitem}
\providecommand{\algorithmiccomment}[1]{\hfill$\triangleright$\ \textit{#1}}

\usepackage[accepted]{icml2026}
\usepackage{amsmath}
\usepackage{mathtools}
\usepackage{amsthm}
\usepackage{siunitx}

\usepackage[capitalize,noabbrev]{cleveref}
\usepackage{subcaption}
\usepackage{booktabs}

\theoremstyle{plain}
\newtheorem{theo}{Theorem}[section]

\theoremstyle{definition}

\theoremstyle{remark}
\newtheorem{remark}[theo]{Remark}

\usepackage[textsize=tiny]{todonotes}

\icmltitlerunning{Amortized Acquisition Optimization for Bayesian Optimization with Variational Mutual Information}

\begin{document}

\twocolumn[
  \icmltitle{Amortized Acquisition Optimization for Bayesian Optimization with Variational Mutual Information}

  \begin{icmlauthorlist}
    \icmlauthor{Farhad Mirkarimi}{yyy}
  \end{icmlauthorlist}

  \icmlaffiliation{yyy}{McGill University, Montreal, Quebec, Canada}
  \icmlkeywords{Bayesian Optimization, Mutual Information, Variational Inference,
                Surrogate Models, Actor-Critic, Machine Learning, ICML}
  \vskip 0.3in
]

\icmlcorrespondingauthor{Farhad Mirkarimi}{farhad.mirkarimi@mail.mcgill.ca}
\printAffiliationsAndNotice{}


\begin{abstract}
Bayesian optimization (BO) of expensive black-box functions is
traditionally addressed with Gaussian processes (GPs), which scale
cubically with observations, or Bayesian neural networks (BNNs), which
incur costly posterior sampling and inner-loop acquisition optimization.
We propose {VBO-MI} (Variational Bayesian Optimization with
Mutual Information), a fully gradient-based BO framework that requires no explicit GP prior or fixed parametric posterior family over objective function and treats it as a strict black box.
An actor-critic architecture pairs an action-net with a variational
critic that estimates information gain, eliminating the acquisition
optimization bottleneck and achieving up to $10^{2}\times$ fewer FLOPs
than BNN-BO baselines.
A lightweight surrogate network further reduces real function queries to
one batch per iteration.
We establish consistency guarantees and evaluate VBO-MI on synthetic
benchmarks (Ackley, Levy, Griewank) and real-world tasks (Rover
Trajectory, Lunar Lander, Pest Control), demonstrating competitive or
superior performance over the baselines.
\end{abstract}
\section{Introduction}
\label{submission}

Bayesian optimization (BO) has matured into a de-facto tool for expensive
global black-box optimization with applications in hyperparameter tuning,
materials design, and experimental
sciences~\cite{frazier2018tutorialbayesianoptimization,garnett_bayesoptbook_2023}.
In Bayesian Optimization, the algorithm treats $f$ as a black-box function
and aims to optimize it sequentially under a finite budget $T$---with
access to only $T$ evaluations (input--output pairs) of the
function~\cite{271617}.
At each evaluation, the algorithm builds a probabilistic model of $f$
using the observed input--output pairs, and uses this model to guide the
selection of the next evaluation points by maximizing an acquisition
function that trades off exploration and exploitation.

Traditionally, GP assumptions for the surrogate coupled with acquisition
functions such as expected improvement~\cite{JMLR:v13:hennig12a,Hennig_Osborne_Kersting_2022,ziqi}
and UCB~\cite{srinivas2010gaussian} have driven significant
progress~\cite{pmlr-v162-rudner22a}.
However, GPs require inverting a covariance matrix (cubic scaling in
observations) and become computationally prohibitive for large datasets.
Although approximate GPs have been proposed~\cite{ermonbo,pmlr-v206-jimenez23a},
they still face trade-offs between accuracy, local fidelity, and uncertainty
quantification.
Recent works replace GPs with BNNs~\cite{li2024a,NIPS2016_a96d3afe},
using variational inference with Sparse Gaussian
assumptions~\cite{NEURIPS2024_da30215e}, Gamma variational
distributions~\cite{cheng2025a}, or
HMC~\cite{li2024a}\nocite{ziqi}.
However, all these methods rely on specific distributional forms for
posterior sampling, limiting their capability in complex,
high-dimensional settings, and incurring significant computational
cost~\cite{cheng2025a,li2024a}.

To address these limitations, this work develops a framework that models
$f$ without assuming a specific parametric family.
By leveraging a flexible variational representation, the proposed approach
enables more expressive uncertainty modeling and improved exploration.
Our method enables full gradient flow through the internal optimization
process while maintaining strict zeroth-order access to $f$.
We leverage recent advances in variational mutual information
estimation~\cite{song2020understanding,pmlr-v80-belghazi18a} to design a
novel acquisition function approximated directly from data samples, jointly
trained with an action-net that explores the input space.
Our main contributions are:
\begin{enumerate}[leftmargin=*,topsep=1pt,itemsep=0pt]
\item We propose a general BO framework that does not assume any specific variational distribution
      form for the surrogate, allowing greater flexibility in modeling
      complex objective functions.
\item A novel variational acquisition function derived from a variational
      MI formulation, providing a principled exploration--exploitation
      trade-off.
\item A lightweight surrogate network that replaces real oracle calls
      inside the inner loops, reducing oracle queries to one batch per
      iteration, and demonstrate its less computational complexity compared to BNN based methods.
\end{enumerate}

\paragraph{Notation.}
Random vectors are denoted by bold lowercase symbols
(e.g., $\mathbf{x}_t \in \mathbb{R}^d$).
The sequence up to step $t$ is $\mathbf{X}_{t} = [\mathbf{x}_1, \dots,
\mathbf{x}_t]$.
A realization uses superscript $(i)$, so $\mathbf{x}_t^{(i)} \in
\mathbb{R}^d$.
Calligraphic letters (e.g.\ $\mathcal{D}$) denote sets.
$I(\mathbf{x};y)$ denotes mutual information between a vector and scaler random variables;
$\mathrm{KL}(P\|Q)$ denotes KL divergence between two probability measures $P$, and $Q$.

\section{Related Work}

\paragraph{Scalable Bayesian optimization.}
GPs remain the canonical choice for BO due to their well-calibrated
uncertainty~\cite{srinivas2010gaussian}, but their $\mathcal{O}(T^3)$
complexity has spurred scalable alternatives: Vecchia
approximations~\cite{pmlr-v206-jimenez23a}, Trust-Region
BO~\cite{NEURIPS2019_6c990b7a}, and Focalized Sparse
GPs~\cite{NEURIPS2024_da30215e}.
While effective, these methods rely on stationary kernel assumptions that
struggle with complex, non-linear landscapes.

\paragraph{Variational and Sampling-Based BNNs.}
BNNs offer superior scalability and representation power.
HMC~\cite{NIPS2016_a96d3afe,li2024a} provides high-quality posteriors at
prohibitive cost.
Variational Inference via VBLL~\cite{harrison2024vbll} and
PFNs~\cite{muller2023pfns} reduce sampling costs but still rely on
parametric posterior families that may cause model misspecification.

\paragraph{Information-Theoretic acquisition functions.}
PES~\cite{pmlr-v37-hernandez-lobatob15} and
MES~\cite{pmlr-v70-wang17e} offer principled non-myopic exploration.
JES~\cite{hvarfner2022joint} models the joint density but reverts to GP
assumptions for tractability.
A recent unified framework~\cite{cheng2025a} introduces acquisition
functions constrained to specific variational families such as Gamma.
Closest to our work, \citet{pmlr-v267-ishikura25a} propose a variational
lower bound for entropy search, but remain rooted in the GP framework.
In contrast, our actor-critic formulation makes no GP assumptions on $f$
and eliminates the inner-loop optimization bottleneck entirely.

\section{Problem Setting and Background}
\label{sec:background}

\subsection{Problem Setting}

We consider maximizing an unknown scalar-valued objective
$f:\mathcal{D}\to\mathbb{R}$ over compact $\mathcal{D}\subseteq\mathbb{R}^d$:
\begin{equation}
    \mathbf{x}^{*} = \arg\max_{\mathbf{x}\in\mathcal{D}} f(\mathbf{x}).
    \label{eq:bo_objective}
\end{equation}
The function is a strict black box: no analytical form or gradients are
available.
At iteration~$t$, the optimizer selects $\mathbf{x}_t$ and receives noisy
observation $y_t = f(\mathbf{x}_t) + \varepsilon_t$, where $\varepsilon_t$
is independent zero-mean noise.
We denote $N = T \cdot B$ the total function evaluations, with each of
$T$ main iterations querying one batch of $B$
points.\footnote{We also perform $W$ warm-up iterations; see
Algorithm~\ref{alg:vbo}.}

\subsection{Bayesian Optimization and Information Gain}

BO maintains a probabilistic surrogate over $f$ and uses an acquisition
function $\alpha(\mathbf{x}\mid\mathbf{X}_t,\mathbf{Y}_t)$ to select the
next input.
The information gain~\cite{srinivas2010gaussian} quantifies information
about $f$ from observations $\mathbf{Y}_T$:
\begin{equation}\label{eq:0}
    I(\mathbf{Y}_{T};f) = H(\mathbf{Y}_{{T}}) - H(\mathbf{Y}_{{T}}\mid f).
\end{equation}
Under a GP assumption, sequentially maximising this gives the GP-UCB
rule~\cite{srinivas2010gaussian}:
\begin{equation}\label{eq:3}
    \mathbf{x}_{t}=\arg\max_{\mathbf{x}\in\mathcal{D}}
    \sqrt{\beta}\,\sigma_{t-1}(\mathbf{x})+ \mu_{t-1}(\mathbf{x}),
\end{equation}
where $\mu_{t-1}$ and $\sigma_{t-1}$ are the GP posterior mean and standard
deviation, and $\beta$ controls the exploration--exploitation trade-off.

\subsection{Variational Mutual Information}
\label{sec:vmi}

To incorporate Mutual Information (MI) into our optimization strategy without relying on restrictive assumptions (e.g., Gaussianity), we employ a variational lower bound based on the Donsker--Varadhan (DV) representation~\cite{pmlr-v80-belghazi18a,https://doi.org/10.1002/cpa.3160280102}~\nocite{9517832,10075335} of the Kullback--Leibler (KL) divergence.

Let $P$ and $Q$ be two probability measures over the same measurable space. The DV representation expresses the KL divergence as
\begin{flalign}\label{Fdiv}
D_{\mathrm{KL}}(P \,\|\, Q) = \sup_{T_{c}} \, \mathbb{E}_{P}[T_{c}] - \log \mathbb{E}_{Q}[e^{T_{c}}], 
\end{flalign}
where the supremum is taken over all measurable functions \(T_{c}\) such that the expectations are finite~\cite{pmlr-v80-belghazi18a,https://doi.org/10.1002/cpa.3160280102}. Now, consider random vector $\mathbf{x}$ and random variable $y$ with joint distribution $P_{\mathbf{x}, y}$, and let $Q = P_{\mathbf{x}}\times P_{{y}}$ denote the product of marginals. Then, the MI between $\mathbf{X}$ and ${y}$ can be written as
\begin{flalign}
&I(\mathbf{x}; {y}) = D_{\mathrm{KL}}\!\big(P_{\mathbf{x},{y}} \,\|\, P_{\mathbf{x} \times {y}}\big) 
=\\& \sup_{T \in \mathcal{T}} \, \mathbb{E}_{P_{\mathbf{x},{y}}}[T(\mathbf{x},{y})] - \log \mathbb{E}_{P_{\mathbf{x}}\times P_{{y}}}[e^{T(\mathbf{x},{y})}]. \label{MIbound}
\end{flalign}
In practice, we parameterize $D(\mathbf{x}, {y})$ using a Neural Network $D_\theta(\mathbf{x}, {y})$ with parameters $\theta$, and estimate the expectations via sampling. This yields the following variational lower bound:
\begin{equation}
I(\mathbf{x}; {y}) \geq \max_{\theta \in \Theta} \, \mathbb{E}_{P_{\mathbf{x},{y}}}[D_\theta(\mathbf{x},{y})] - \log \mathbb{E}_{P_{\mathbf{x}} \times P_{{y}}}[e^{D_\theta(\mathbf{x},{y})}].
\end{equation}
Given samples of distributions $\{(\mathbf{x}^{(i)}, {y}^{(i)})\}_{n=1}^N$, the Neural Network $T_{\theta}$ is trained by minimizing the following loss function:

\begin{flalign}\label{miestimate}
L_{D_{\theta}} = - \frac{1}{B} \sum_{i=1}^B D_\theta(\mathbf{x}^{(i)}, y^{(i)}) + \log\left( \frac{1}{B^2} \sum_{i=1}^B \sum_{j=1}^B e^{D_\theta(\mathbf{x}^{(i)}, y^{(j)})} \right),
\end{flalign}
\normalsize
where \(B\) denotes the minibatch size. This formulation provides a scalable, sample-based estimator of MI, and integrate into our Bayesian Optimization framework to guide exploration.

\begin{figure*}[t]
\centering
\resizebox{1.0\textwidth}{!}{%
\tikzset{every picture/.style={line width=0.75pt}}
\begin{tikzpicture}[x=0.75pt,y=0.75pt,yscale=-1,xscale=1]


\draw [color={rgb,255:red,52;green,152;blue,219},draw opacity=0.30,
       fill={rgb,255:red,52;green,152;blue,219},fill opacity=0.13]
  (0,90) rectangle (310,455);

\draw [color={rgb,255:red,26;green,188;blue,156},draw opacity=0.28,
       fill={rgb,255:red,26;green,188;blue,156},fill opacity=0.10]
  (308,90) rectangle (608,455);

\draw [color={rgb,255:red,192;green,57;blue,43},draw opacity=0.22,
       fill={rgb,255:red,192;green,57;blue,43},fill opacity=0.07]
  (606,90) rectangle (1100,455);

\draw (150,108) node {\Large\textbf{Action-Net} $E_\phi$};
\draw (458,108) node {\Large\textbf{Surrogate} $\hat{f}_\psi$};
\draw (878,108) node {\Large\textbf{Helper} $D_\theta$};


\draw [fill={rgb,255:red,189;green,195;blue,199},fill opacity=0.65]
  (8,196) rectangle (82,240);
\draw (45,210) node {\large Seed};
\draw (45,228) node {$\mathbf{s}_t \!\sim\! \mathcal{N}(\mathbf{0},\mathbf{I})$};

\draw [fill={rgb,255:red,52;green,152;blue,219},fill opacity=0.40]
  (94,158) -- (248,178) -- (248,262) -- (94,282) -- cycle;
\draw (165,206) node {\large Action-net};
\draw (165,225) node {$E_{\phi}(\mathbf{s}_t)$};

\draw [line width=1.5,->] (82,218) -- (93,218);

\draw [fill={rgb,255:red,184;green,233;blue,134},fill opacity=0.55]
  (258,200) -- (300,200) -- (300,236) -- (258,236) -- cycle;
\draw (279,210) node {\small Input};
\draw (279,225) node {$\mathbf{x}_t^{(i)}$};

\draw [line width=1.5,->] (248,218) -- (257,218);

\draw [fill={rgb,255:red,26;green,188;blue,156},fill opacity=0.35,
       rounded corners=8pt]
  (320,185) -- (498,185) -- (498,253) -- (320,253) -- cycle;
\draw (409,207) node {\large Surrogate network};
\draw (409,228) node {$\hat{f}_{\psi}(\mathbf{x}_t^{(i)})\!=\!\hat{y}_t^{(i)}$};

\draw [line width=1.5,->] (300,218) -- (319,218);


\draw [fill={rgb,255:red,184;green,233;blue,134},fill opacity=0.55]
  (508,200) -- (560,200) -- (560,236) -- (508,236) -- cycle;
\draw (534,210) node {\small Pred.};
\draw (534,225) node {$\hat{y}_t^{(i)}$};

\draw [line width=1.5,->] (498,218) -- (507,218);

\draw [fill={rgb,255:red,192;green,57;blue,43},fill opacity=0.28,
       rounded corners=8pt]
  (625,175) -- (910,175) -- (910,265) -- (625,265) -- cycle;
\draw (737,207) node {\large LSTM + MLP};
\draw (737,230) node {$D_{\theta}\!\left(\mathbf{x}_t^{(i)},\,\hat{y}_t^{(i)}\right)$};

\draw [line width=1.5,->] (560,218) -- (624,218);

\draw [fill={rgb,255:red,192;green,57;blue,43},fill opacity=0.22,
       rounded corners=5pt]
  (925,192) -- (1070,192) -- (1070,245) -- (925,245) -- cycle;
\draw (997,210) node {$L_{D_\theta}$};
\draw (997,230) node {\small (DV variational bound)};

\draw [line width=1.5,->] (910,218) -- (924,218);

\draw [line width=1.3,->,dashed,
       color={rgb,255:red,192;green,57;blue,43},draw opacity=0.90]
  (947,192) -- (947,143) -- (717,143) -- (717,175);
\draw (840,133) node [color={rgb,255:red,192;green,57;blue,43}]
  {$\nabla_\theta L_{D_\theta}$};

\draw [fill={rgb,255:red,26;green,188;blue,156},fill opacity=0.25]
  (458,128) -- (518,148) -- (458,168) -- (398,148) -- cycle;
\draw (458,148) node {\large $L_{E_\phi}$};

\draw [line width=1.2,->,dashed,draw opacity=0.70]
  (459,185) -- (459,168);
\draw (385,138) node {\small exploit};

\draw [line width=1.2,->,dashed,draw opacity=0.70]
  (695,175) -- (695,148) -- (520,148);
\draw (613,138) node {\small explore ($I_\theta$)};

\draw [line width=1.4,->,dashed,
       color={rgb,255:red,41;green,128;blue,185},draw opacity=0.92]
  (398,148) -- (165,148) -- (165,158);
\draw (278,136) node [color={rgb,255:red,41;green,128;blue,185}]
  {$\nabla_\phi L_{E_\phi}$};


\draw [line width=1.0,dashed,
       color={rgb,255:red,127;green,140;blue,141},rounded corners=4pt]
  (252,285) -- (1090,285) -- (1090,440) -- (252,440) -- cycle;
\draw (511,300) node [color={rgb,255:red,127;green,140;blue,141}]
  {\small\textit{Single real oracle query per BO iteration}};

\draw [line width=1.5,->] (279,236) -- (279,325);
\draw [line width=1.5,->] (279,325) -- (318,325);

\draw [fill={rgb,255:red,230;green,126;blue,34},fill opacity=0.52,
       rounded corners=7pt]
  (320,310) -- (498,310) -- (498,360) -- (320,360) -- cycle;
\draw (409,326) node {\large\textbf{Black-box oracle} $f$};
\draw (409,344) node {$y_t^{(i)} \!=\! f(\mathbf{x}_t^{(i)}) \!+\! \epsilon^{(i)}$};

\draw [line width=1.5,->] (498,334) -- (548,334);
\draw (560,325) node {\small $y_t^{(i)}$};

\draw [line width=1.5,->] (576,334) -- (618,334);

\draw [fill={rgb,255:red,149;green,165;blue,166},fill opacity=0.55,
       rounded corners=7pt]
  (620,310) -- (870,310) -- (870,360) -- (620,360) -- cycle;
\draw (745,326) node {\large Replay buffer};
\draw (745,345) node {$\mathcal{H} \leftarrow \mathcal{H} \cup \{(\mathbf{x}_t^{(i)}, y_t^{(i)})\}$};

\draw [line width=1.4,->,
       color={rgb,255:red,22;green,160;blue,133}]
  (745,310) -- (745,278) -- (409,278) -- (409,253);
 \draw (588,270) node [align=center,
        color={rgb,255:red,22;green,160;blue,133}]
   {\small train $\hat{f}_\psi$ on $\mathcal{L}_{\mathrm{MSE}}$ ($K_s$ gradient steps)};

\end{tikzpicture}%
}
\caption{%
VBO-MI framework. 
An \emph{action-net} $E_\phi$ maps random seeds $\mathbf{s}_t$ to candidate inputs $\mathbf{x}_t^{(i)}$.
A \emph{surrogate} $\hat{f}_\psi$ predicts rewards $\hat{y}_t^{(i)}$, enabling
oracle-free updates of the helper $D_\theta$ ($K_a$ steps) and action-net ($K_b$ steps)
via the acquisition loss $L_{E_\phi}$, which balances exploitation (surrogate mean)
and exploration (variational MI estimate $I_\theta$).
One real oracle query per iteration yields $y_t^{(i)}$, which replenishes the replay
buffer $\mathcal{H}$ used to retrain $\hat{f}_\psi$ via MSE ($K_s$ steps).
Dashed arrows indicate gradient flows.%
}
\label{fig:vertical_rotated}
\end{figure*}
\section{VBO-MI: Proposed Framework}
\label{secvbo}
\subsection{Exploration and Mutual Information Gain Estimation}
Using results of previous section, our goal here is to find a variational bound for exploring the function $f$.
Based on the defined setting, at time step $t$ we have access to $\mathbf{X}_{t}=[\mathbf{x}_{1},...,\mathbf{x}_{t}],$ of current chosen actions. We also, assume that these actions are chosen according to $P_{\mathbf{X}_{t}}$. Moreover, corresponding observations are $\mathbf{Y}_{t}=[{y}_{1},..,{y}_{t}]$, where ${y}_{t}=f(\mathbf{x}_{t})+{\epsilon_{t}}$, and
$\epsilon_{t}$ is zero mean, independent, identically distributed (i.i.d.) noise.  Our goal in this section is to find a lower bound on the information gain quantity which corresponds to the exploration term in ~\eqref{eq:3}, which is:
\begin{equation}\label{eq:MI-exp}
    I_{G}=I(f(\mathbf{X}_t),\mathbf{Y}_{t}),
\end{equation}
where we have\footnote{Depending on its argument, $f(\mathbf{x}_{t})$ represents a scalar value, or a vector, $f(\mathbf{X}_{t})$.} $\mathbf{Y}_{t}=f(\mathbf{X}_t)+{\mathcal{E}}_{t}$, $\mathcal{E}_{t}=[\epsilon_{i}]_{i=1}^{t}$, and $f(\mathbf{X}_t)=[f(\mathbf{x}_{i})]_{i=1}^{t}$.\footnote{ Note that here we don't have a Gaussian assumption (for $f$ or noise), and therefore we cannot use the closed form for Gaussian distribution entropy)}
To find a simple bound on this quantity using the bound of~\cite{pmlr-v80-belghazi18a}, and ~\eqref{MIbound}, we first note that we have the following Markov chain~\cite{cover2012elements}:
\begin{equation}\label{markov0}
    \mathbf{X}_{t}\longleftrightarrow f(\mathbf{X}_{t})\longleftrightarrow\mathbf{Y}_{t}\longleftrightarrow y_t.
\end{equation}
Based on the data-processing inequality~\cite{cover2012elements} we have 
\begin{equation}\label{IG-f-1}
    I(\mathbf{X}_{t};{y}_{t})\leq I_{G}=I(f(\mathbf{X}_t),\mathbf{Y}_{t}).
\end{equation}
By expanding the mutual information via chain rule we have:
\begin{flalign}
   & I(\mathbf{X}_{t};y_t)=I(\mathbf{x}_{t};y_t)+I(\mathbf{x}_{t-1};y_{t}\mid \mathbf{x}_{t})\\&+I(\mathbf{x}_{t-2};y_t\mid \mathbf{x}_{t},\mathbf{x}_{t-1})\\&+\dots+I(\mathbf{x}_{1};y_{t}\mid \mathbf{x_{t}},\dots \mathbf{x}_{2}),
\end{flalign}
except for the first term, all the conditional terms are zero due to Markov properties~\cite{cover2012elements}. On the other hand, by using the variational representation in ~\eqref{Fdiv}:
\begin{flalign}\label{eq:IG-f}
  & I_{\theta}(\mathbf{x}_{t},{y}_{t})= \sup_{\theta\in\Theta}\mathbb{E}_{P_{\mathbf{x}_{t},{y}_{t}}}[D_{\theta}(\mathbf{x}_{t},{y}_{t})]+\nonumber\\&\log (\mathbb{E}_{ P_{\mathbf{x}_{t}}\times P_{{y}_{t}}}[e^{D_{\theta}(\mathbf{x}_{t},{y}_{t})}]).
\end{flalign}
By noting that $I_{\theta}(\mathbf{x}_{t},{y}_{t})\leq I(\mathbf{x}_{t},{y}_{t}),$ and combining ~\eqref{IG-f-1} and ~\eqref{eq:IG-f} we have the following bound on the exploration term, or IG bound in ~\eqref{eq:3}:
\begin{equation}\label{eq:IG-f+1}
    \mathbb{E}_{P_{\mathbf{x}_{t},{y}_{t}}}[D_{\theta}(\mathbf{x}_{t},{y}_{t})]+\log (\mathbb{E}_{ P_{\mathbf{x}_{t}}\times P_{{y}_{t}}}[e^{D_{\theta}(\mathbf{x}_{t},{y}_{t})}]) \leq I_{G}.
\end{equation}
We choose $y_{t}$, the current observation, rather than the full history $\mathbf{Y}_{t}$: by the Markov chain in~\eqref{markov0}, $y_{t}$ captures the \emph{marginal} information gained at step $t$ given the already-observed history, avoiding double-counting of information already incorporated in previous iterations. This choice allows the 
exploration signal to remain focused on the incremental uncertainty 
reduction provided by the newest query, rather than considering redundant historical observations which would also increase the complexity of evaluating the mutual information bound.
\subsection{A Data-Driven Exploitation term}
To take into account the exploitation capability of our algorithm, we also estimate exploitation term using a data-driven approach. The goal is use these results in our acquisition function (loss of a neural network) to choose the best actions to approximate and maximize $f$. In other word, our loss function should simultaneously consider the balance of exploration and exploitation in searching the input space. In the previous subsection, we have seen how the information gain quantity can be leveraged to consider the exploration of the input space. To account for the exploitation capability of our algorithm, we need to consider finding the maximum of $f$ by utilizing our current knowledge. A straightforward approach to find the maximum is to consider the expectation of the function $f$ conditioned on previous points (posterior of $f$), e.g:
\begin{equation}
    \mathbb{E}[f\mid y_{t-1},..,y_{1},\mathbf{x}_{t-1},..,\mathbf{x}_{1}].
\end{equation}
Furthermore, we note that by assuming that $f$ and noise are independent, and noise has a zero mean, we have:
\begin{flalign}
   &\mathbb{E}[f\mid y_{t-1},..,y_{1},\mathbf{x}_{t-1},..,\mathbf{x}_{1}]=\nonumber \\&\mathbb{E}[f+\epsilon_{t}\mid y_{t-1},..,y_{1},\mathbf{x}_{t-1},..,\mathbf{x}_{1}]= \\&\mathbb{E}[y_{t}\mid y_{t-1},..,y_{1},\mathbf{x}_{t-1},..,\mathbf{x}_{1}].\label{explt1}
\end{flalign}
In practice, computing $\mathbb{E}[y_t \mid y_{t-1},\dots,y_1,
\mathbf{x}_{t-1},\dots,\mathbf{x}_1]$ analytically requires a 
parametric surrogate such as a Gaussian Process, which reintroduces 
the assumptions we seek to avoid.  Instead, we approximate this 
conditional expectation directly from data using a neural network 
surrogate $\hat{f}_\psi : \mathcal{D} \to \mathbb{R}$, trained by 
minimizing the mean squared error on all past observations in replay buffer 
$\mathcal{H} = \{(\mathbf{x}^{(i)}_{t'}, y^{(i)}_{t'})\},\, t^{'}\leq t$:
\begin{flalign}
   & \mathcal{L}_{\mathrm{MSE}}(\psi) 
    = \frac{1}{B}\sum_{i=1}^{B}
      \bigl(\hat{f}_\psi(\mathbf{x}^{(i)}) - y^{(i)}\bigr)^2,
      \\
      &\{(\mathbf{x}^{(i)}, y^{(i)})\}_{i=1}^{B} 
      \overset{\mathrm{i.i.d.}}{\sim} \mathrm{Uniform}(\mathcal{H}).
\end{flalign}
 The trained surrogate $\hat{f}_\psi(\mathbf{x})$ then serves as a 
point estimate of $\mathbb{E}[y_t \mid \mathcal{H}]$, which we call it $\hat{y}_{t}$ at any proposed 
action $\mathbf{x}$, and its gradient $\nabla_\mathbf{x}\hat{f}_\psi$ 
provides a differentiable exploitation signal for updating $E_\phi$ 
without any parametric assumption on $f$ and without querying the 
black-box function.
\subsection{Variational Formulation of BO}

Our first observation is that the GP-UCB policy~\eqref{eq:3} implicitly
depends on the MI between $f$ and the observations $\mathbf{y}_t$
conditioned on selected actions.
Based on the previous sections proposed exploration and exploitation terms, we formulate the BO problem as:
\begin{equation}\label{mnet}
    \phi^{*} = \operatorname*{arg\,max}_{\phi}\min_{\theta}\; g(\phi,\theta),
    \qquad
    \mathbf{x}_{t} = E_{\phi^{*}}(\mathbf{s}_{t}),
\end{equation}
where $E_{\phi}$ (action-net) and $D_{\theta}$ (helper) are two neural
networks trained in an alternating actor-critic fashion. In a high-level view the action-net search in the input space to find optimal inputs and helper network computes mutual information and contributes to computation of our acquisition function ($g$) to guide action-net.
To the best of our knowledge, this is the first fully gradient-based BO
framework based on variational formulations without GP or specific
variational family assumptions.

\subsection{Phase~I: Surrogate Training and Helper Update}

At each BO iteration $t$, the algorithm maintains a replay buffer
$\mathcal{H} = \{(\mathbf{x}^{(i)}_{t'}, y^{(i)}_{t'})\}$ of all past
observations.
A lightweight surrogate $\hat{f}_{\psi}:\mathcal{D}\to\mathbb{R}$
approximates the conditional expectation $\mathbb{E}[y_t\mid\mathcal{H}]$
by minimizing:
\begin{flalign}\label{eq:surr-loss}
    &\mathcal{L}_{\mathrm{MSE}}(\psi)
    = \frac{1}{B}\sum_{i=1}^{B}
      \bigl(\hat{f}_{\psi}(\mathbf{x}^{(i)}) - y^{(i)}\bigr)^{2},\\
    &\{(\mathbf{x}^{(i)},y^{(i)})\}
    \!\overset{\mathrm{i.i.d.}}{\sim}\!
    \mathrm{Uniform}(\mathcal{H}).
\end{flalign}
Because $\hat{f}_\psi$ is trained exclusively on observed input--output
pairs and its gradient flows through the surrogate network (never through
$f$), the black-box condition is never violated.
Surrogate predictions $\hat{y}^{(i)}_t = \hat{f}_{\psi}(\mathbf{x}^{(i)}_t)$
replace real observations in the helper's loss, allowing $D_\theta$ to
update without additional queries to $f$:
\begin{align}\label{eq:-mi}
    L_{D_{\theta}}(\mathbf{x}_{t}, \hat{y}_{t})
    &= -\mathbb{E}_{P_{\mathbf{x}_{t}, \hat{y}_{t}}}
             [D_{\theta}(\mathbf{x}_{t}, \hat{y}_{t})] \nonumber \\
    &\quad+ \log\mathbb{E}_{P_{\mathbf{x}_{t}}\times P_{\hat{y}_{t}}}
             [e^{D_{\theta}(\mathbf{x}_{t}, \hat{y}_{t})}],
\end{align}
with $\phi$ and $\psi$ frozen.
This phase runs for $K_a$ gradient steps.

\begin{remark}
Because $\hat{f}_\psi$ is trained by minimizing the MSE loss over the
full replay buffer $\mathcal{H}$, it summarizes all past observations
into a single differentiable model.
This allows the helper loss~\eqref{eq:-mi} to operate on only the
current-batch pair $(\mathbf{x}_t^{(i)}, \hat{y}_t^{(i)})$ rather than
variable-length history sequences, removing the need for the helper to
process growing observation histories directly and keeping the per-iteration
computation cost constant in~$t$.
\end{remark}

\subsection{Phase~II: Action-Net Update}

With the helper frozen, the action-net $E_{\phi}$ is trained to jointly
maximise exploitation of the surrogate's predicted reward and exploration
measured by the MI estimate from Phase~I.
The combined loss (to be minimized) is:
\begin{equation}
\begin{aligned}
    L_{E_{\phi}}(\mathbf{x}_{t},\hat{y}_{t})
    &= -g({\phi,\theta})
    = -\hat{y}_{t}
    -\sqrt{\beta}\,\mathbb{E}_{P_{\mathbf{x}_{t},\hat{y}_{t}}}
        [D_{\theta}(\mathbf{x}_{t},\hat{y}_{t})] \\
    &\quad+\sqrt{\beta}\log\mathbb{E}_{P_{\mathbf{x}_{t}}\times P_{\hat{y}_{t}}}
        [e^{D_{\theta}(\mathbf{x}_{t},\hat{y}_{t})}],
\end{aligned}
\label{loss-f}
\end{equation}
where $\beta$ controls exploration--exploitation.
Minimizing $L_{E_\phi}$ is equivalent to maximising
$\hat{f}_\psi(\mathbf{x}_t) + \sqrt{\beta}\,I_\theta(\mathbf{x}_t,\hat{y}_t)$.
Implemented as sample averages:
\begin{equation}
\begin{aligned}
    L_{E_{\phi}}
    &= -\frac{1}{B}\sum_{i=1}^{B}\hat{y}^{(i)}_{t}
    +\frac{1}{B}\sum_{i=1}^{B}
        D_{\theta}(\mathbf{x}^{(i)}_{t},\hat{y}^{(i)}_{t}) \\
    &\quad-\log\frac{1}{B^{2}}
        \sum_{i,j}
        e^{D_{\theta}(\mathbf{x}^{(i)}_{t},\hat{y}^{(j)}_{t})},
\end{aligned}
\label{vbomi-loss}
\end{equation}
where $\hat{y}^{(i)}_{t}=\hat{f}_{\psi}(\mathbf{x}^{(i)}_{t})$.
This runs for $K_b$ steps with $D_\theta$ and $\hat{f}_\psi$ frozen.
\begin{remark}
VBO-MI uses gradients to update $\phi$, $\theta$, and $\psi$ via the
variational and MSE losses respectively.
These gradients flow entirely through the neural architectures; $f$
remains a strict black box with no gradients required or used.
\end{remark}

In summary, our framework has three neural network components:
$E_{\phi}$ searches the input space and selects $\mathbf{x}_{t}$;
$D_{\theta}$ estimates information gain (exploration);
$\hat{f}_{\psi}$ approximates the conditional expectation of $f$
(exploitation) (see Fig.\ref{fig:vertical_rotated}).
The algorithm is summarized in Algorithm~\ref{alg:vbo}.
\begin{figure*}[t]
\centering
\begin{subfigure}{\textwidth}
    \centering
    \includegraphics[width=0.7\linewidth]{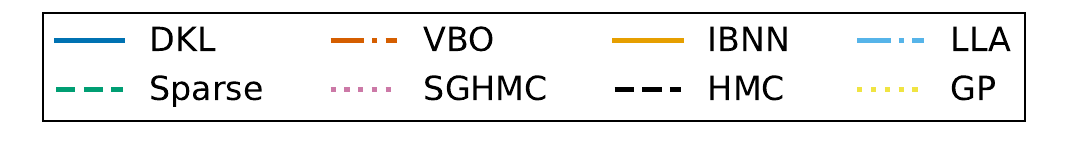}
\end{subfigure}
\vspace{0.1cm}
\begin{subfigure}{0.32\textwidth}
    \centering
    \includegraphics[width=\linewidth]{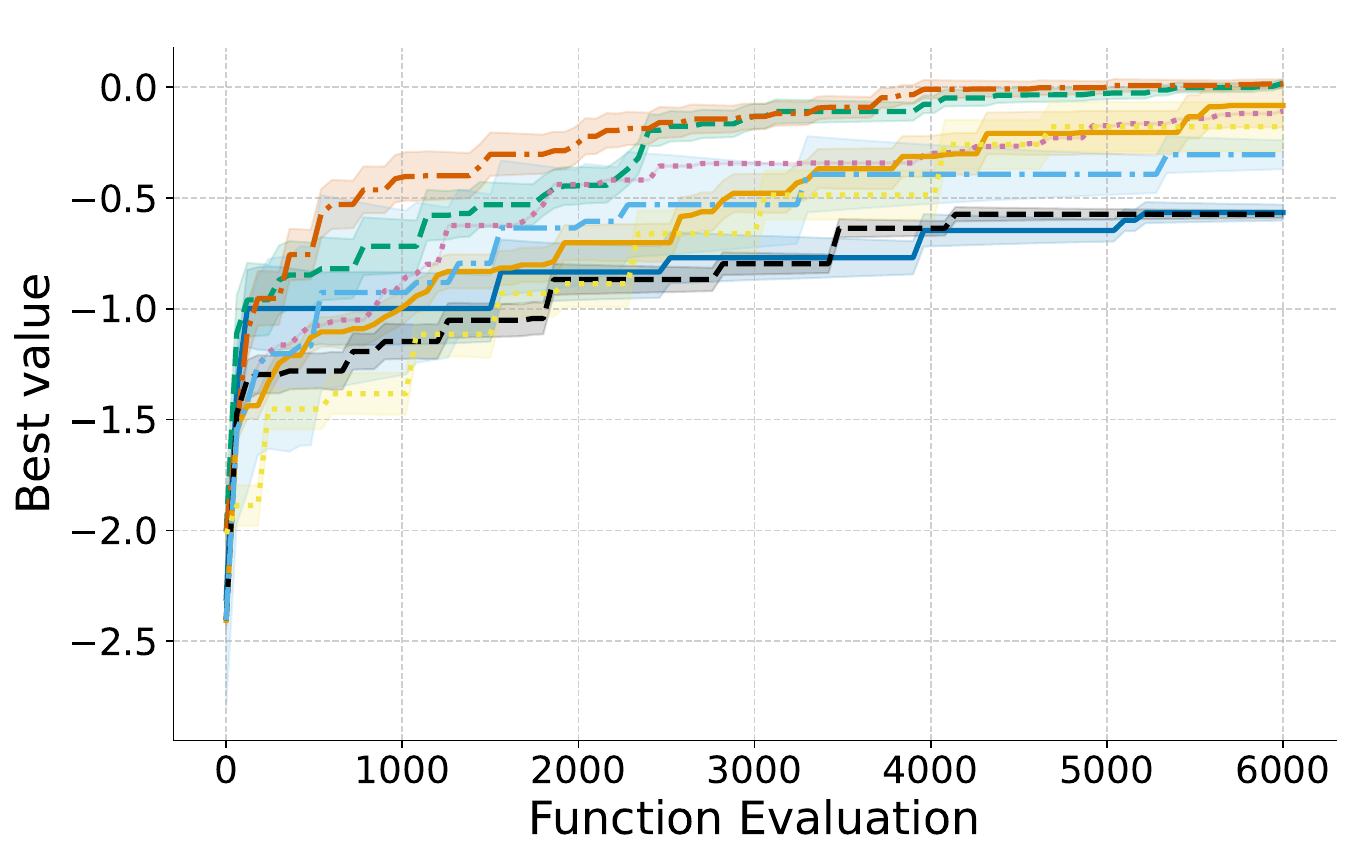}
    \caption{Ackley ($d=10$)}
    \label{fig:ackley}
\end{subfigure}
\hfill
\begin{subfigure}{0.32\textwidth}
    \centering
    \includegraphics[width=\linewidth]{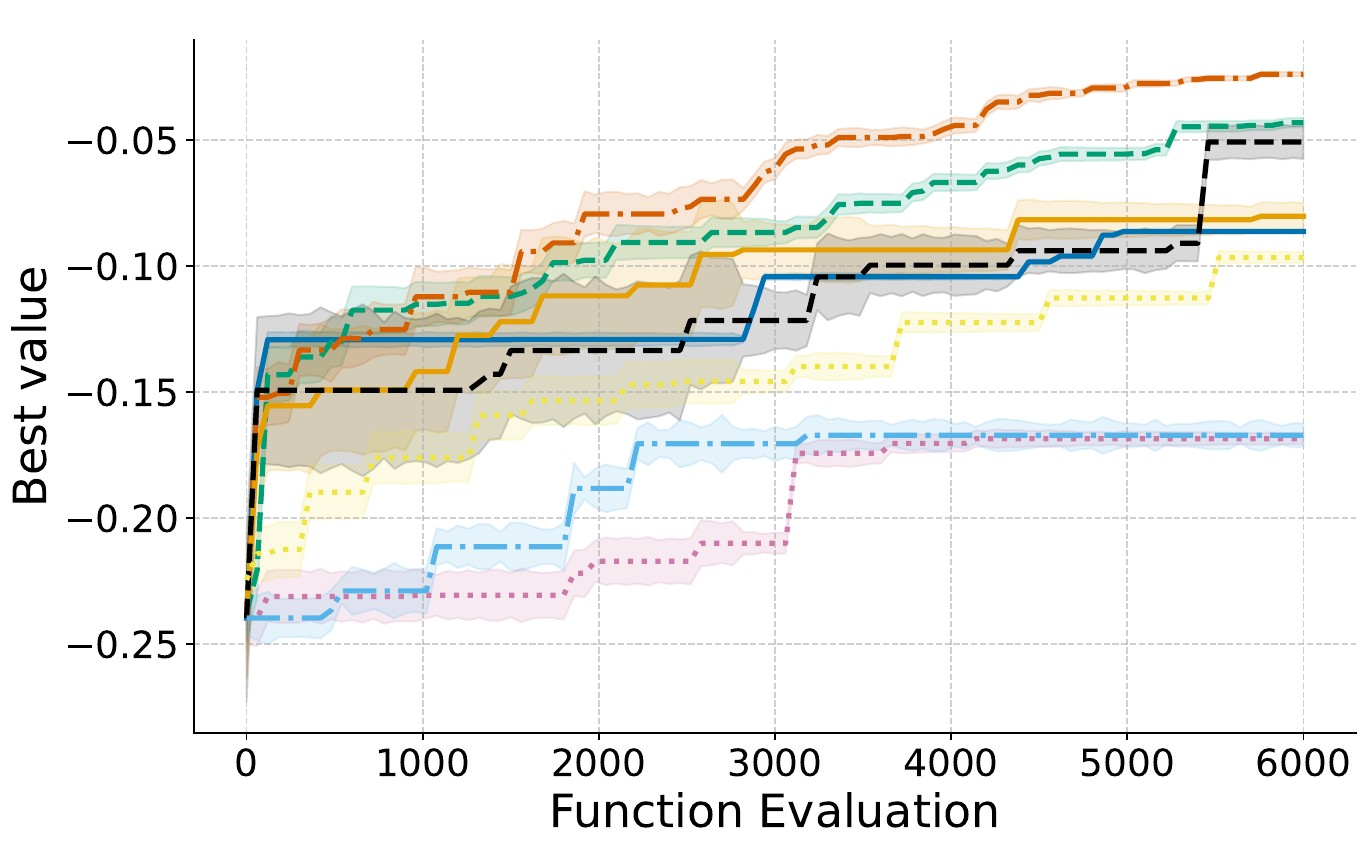}
    \caption{Levy ($d=20$)}
    \label{fig:levy}
\end{subfigure}
\hfill
\begin{subfigure}{0.32\textwidth}
    \centering
    \includegraphics[width=\linewidth]{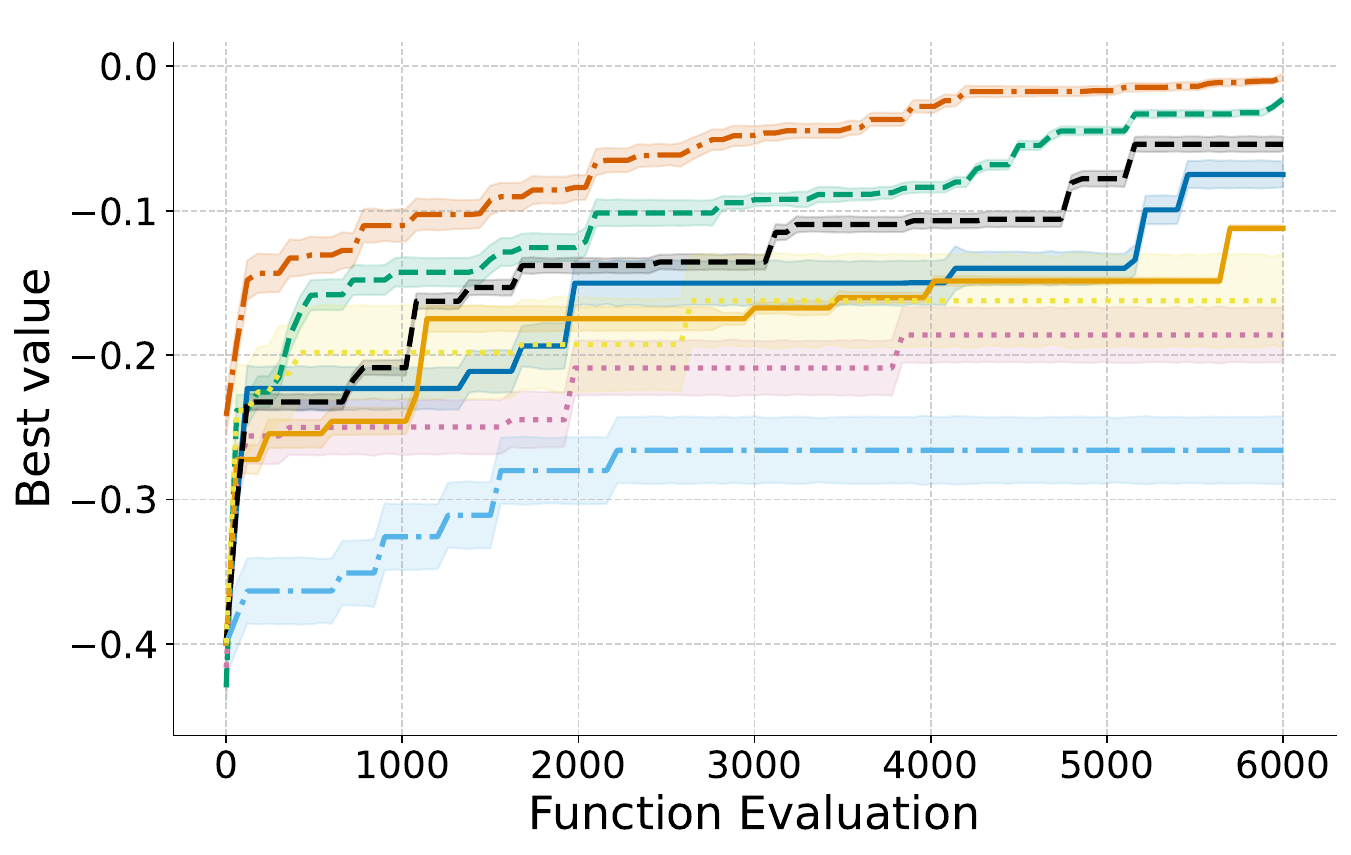}
    \caption{Griewank ($d=20$)}
    \label{fig:griewank}
\end{subfigure}
\caption{Synthetic benchmarks. Global optimum is 0 (dashed line).
         Mean $\pm$ 1 std, 5 trials.}
\label{fig:synthetic}
\end{figure*}

\begin{algorithm}[t]
\caption{VBO-MI}\label{alg:vbo}
\begin{algorithmic}[1]
\STATE \textbf{Input:} Weights of $E_\phi$, $D_\theta$, $\hat{f}_\psi$;\;
       $W$, $T$, $K_a$, $K_b$, $K_s$, $B$, $\beta$.
\STATE \textbf{Output:} $(\phi^*,\theta^*,\psi^*)$; estimated optimum $\mathbf{x}^*$.
\STATE Initialise replay buffer $\mathcal{H}\leftarrow\emptyset$.
\FOR[Warm-up phase]{$k=1,\dots,W$}
  \STATE Sample $\mathbf{x}^{(i)}\!=\!E_\phi(\mathbf{s}_k)$,
         evaluate $y^{(i)}\!=\!f(\mathbf{x}^{(i)})\!+\!\epsilon^{(i)}$,
         store $(\mathbf{x}^{(i)},y^{(i)})$ in $\mathcal{H}$.
  \STATE Update $D_\theta$ on $L_{D_\theta}$ with $E_\phi$ fixed.
\ENDFOR
\FOR[Pre-train surrogate $\hat{f}_\psi$]{$k=1,\dots,200$}
  \STATE Sample mini-batch from $\mathcal{H}$;
         update $\hat{f}_\psi$ on $\mathcal{L}_{\mathrm{MSE}}$.
\ENDFOR
\FOR[Main VBO-MI phase]{$t=1,\dots,T$}
  \STATE Draw $\mathbf{s}_t\!\sim\!\mathcal{N}(\mathbf{0},\mathbf{I})$;\;
         sample $\mathbf{x}^{(i)}_t\!=\!E_\phi(\mathbf{s}_t)$, $i=1,\dots,B$.
  \FOR[Update helper $D_\theta$]{$k=1,\dots,K_a$}
    \STATE $\hat{y}^{(i)}_t\!=\!\hat{f}_\psi(\mathbf{x}^{(i)}_t)$;\;
           update $D_\theta$ on
           $L_{D_\theta}(\mathbf{x}^{(i)}_t,\hat{y}^{(i)}_t)$.
  \ENDFOR
  \FOR{$k=1,\dots,K_s$}
    \STATE Sample mini-batch $\sim\!\mathcal{H}$;\;
           update $\hat{f}_\psi$ on $\mathcal{L}_{\mathrm{MSE}}$.
  \ENDFOR
  \FOR[Update action-net $E_\phi$]{$k=1,\dots,K_b$}
    \STATE Compute $I_\theta$ via frozen $D_\theta$;\;
           update $E_\phi$ on $L_{E_\phi}$~\eqref{vbomi-loss}.
  \ENDFOR
  \STATE Evaluate $y^{(i)}_t\!=\!f(\mathbf{x}^{(i)}_t)\!+\!\epsilon^{(i)}$;\;
         update $\mathcal{H}$. \COMMENT{single real query per iteration}
  \STATE $\mathbf{x}^*_t\!\leftarrow\!
         \operatorname*{arg\,max}_{(\mathbf{x},y)\in\mathcal{H}}\;y$.
\ENDFOR
\STATE \textbf{Return} $(\phi^*,\theta^*,\psi^*)$ and
       $\mathbf{x}^*\!=\!\mathbf{x}^*_T$.
\end{algorithmic}
\end{algorithm}
\label{sec:ablation}
\begin{figure*}[t]
\centering

\begin{subfigure}{0.32\textwidth}
    \centering
    \includegraphics[width=\linewidth]{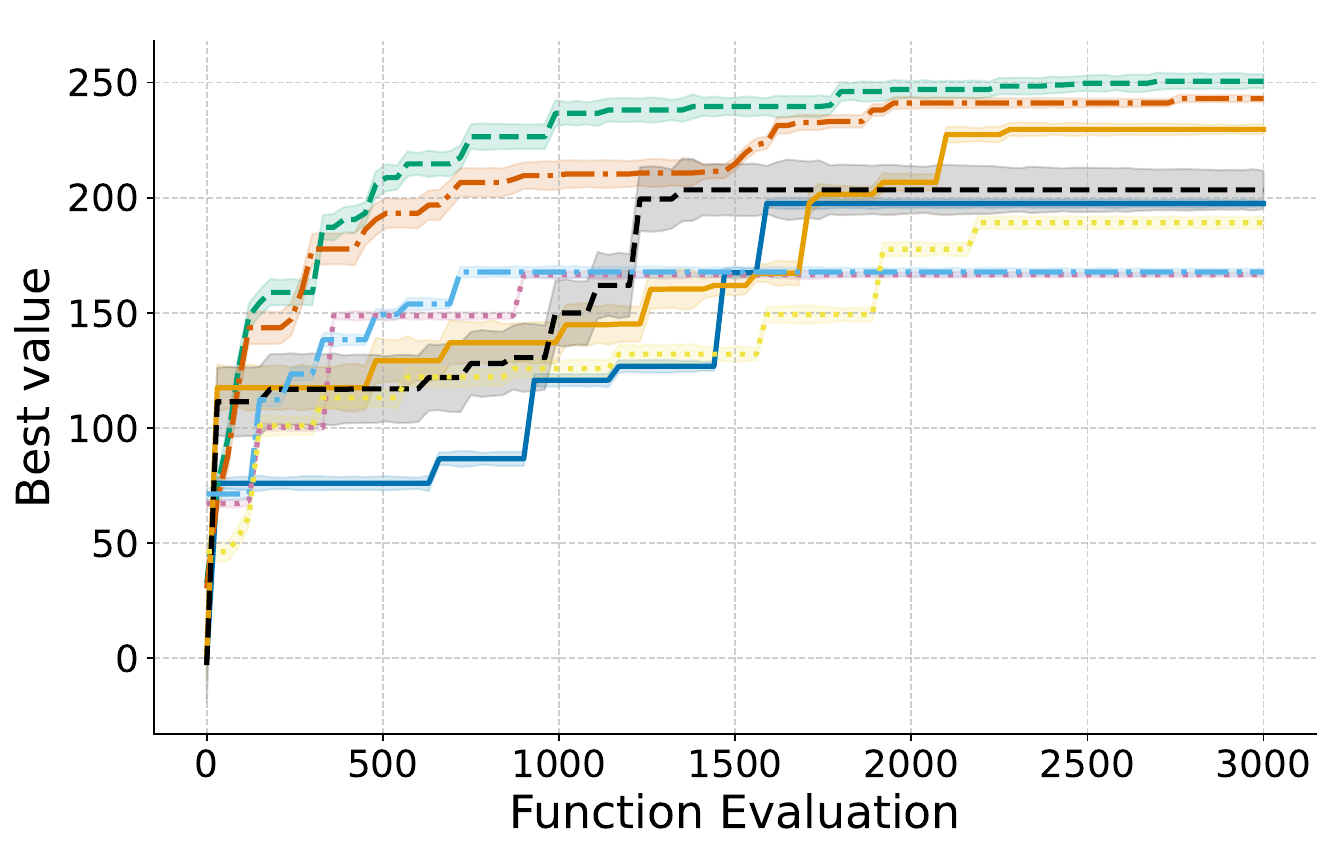}
    \caption{Lunar Lander ($d=12$)}
    \label{fig:lunar}
\end{subfigure}
\hfill
\begin{subfigure}{0.32\textwidth}
    \centering
    \includegraphics[width=\linewidth]{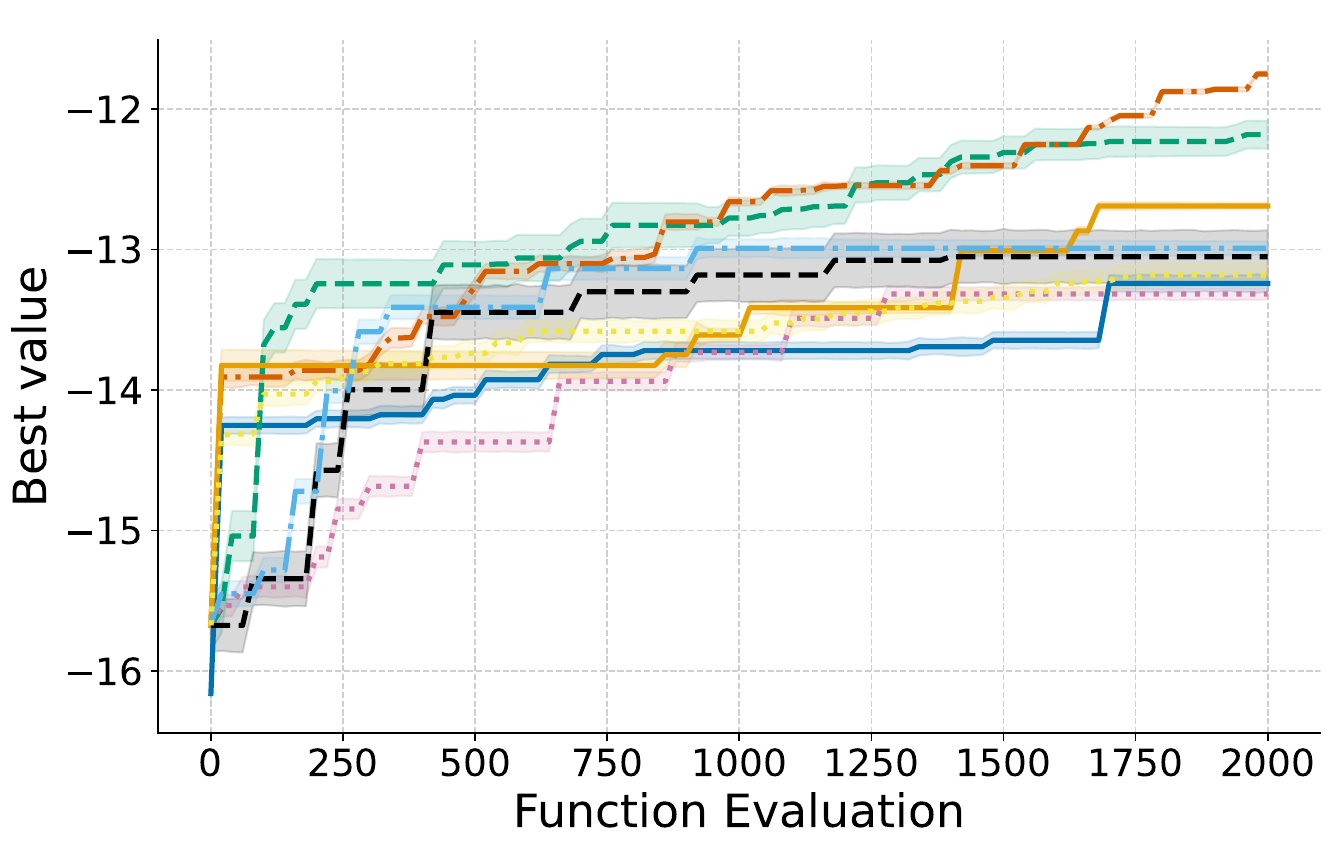}
    \caption{Pest Control ($d=25$)}
    \label{fig:pest}
\end{subfigure}
\hfill
\begin{subfigure}{0.32\textwidth}
    \centering
    \includegraphics[width=\linewidth]{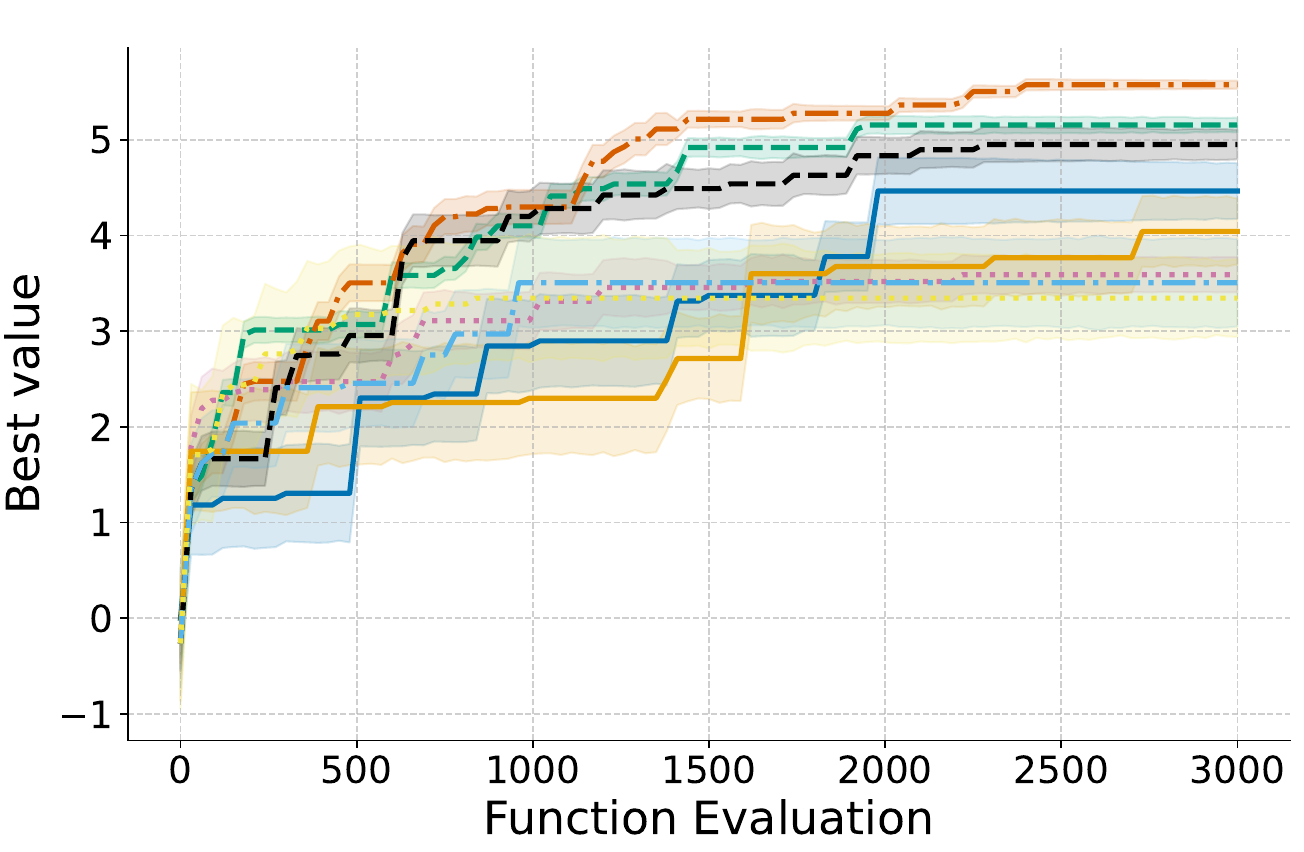}
    \caption{Rover Trajectory ($d=60$)}
    \label{fig:rover}
\end{subfigure}
\caption{Real-world benchmarks. Mean $\pm$ 1 std, 5 trials.}
\label{fig:realworld}
\end{figure*}
\subsection{Consistency Theorem}

\begin{theo}[Consistency of VBO-MI]
\label{thm:consistency}
Let $f:\mathcal{D}\to\mathbb{R}$ be continuous and let
$\widehat{g}(\phi,\theta)$ be the sample-average approximation of
$g(\phi,\theta)$ defined in~\eqref{mnet} and~\eqref{vbomi-loss}.
Assume the DV expectations are finite, the replay buffer $\mathcal{H}$
forms a stationary ergodic sequence, and $E_\phi$, $D_\theta$,
$\hat{f}_\psi$ have sufficient capacity.
Then $\widehat{g}(\phi,\theta)\to g(\phi,\theta)$ uniformly
$\mathbb{P}$-a.s.\ as $B\to\infty$, and for any $\epsilon>0$ there
exists $E_{\hat\phi}$ such that
$\sup_{\mathbf{s}}\|E_{\hat\phi}(\mathbf{s})-\pi^*(\mathbf{s})\|
\leq\epsilon$,
where
$\pi^*(\mathbf{s})=
\operatorname{arg\,max}_{\mathbf{x}\in\mathcal{D}}g_\theta(\mathbf{x})$,
and consequently
$|f(E_{\hat\phi}(\mathbf{s}))-\max_{\mathbf{x}\in\mathcal{D}}f(\mathbf{x})|
\leq L_f\epsilon$
for some modulus of continuity $L_f$.
\end{theo}

\begin{proof}
See Appendix~\ref{theory}.
\end{proof}

\section{Computational Complexity}
\label{sec:complexity}

A key bottleneck in BO is the overhead from surrogate updates,
inner-loop acquisition optimization, and real function evaluations.
VBO-MI matches the standard BO oracle budget of $\mathcal{O}(B)$ queries
per iteration: the surrogate $\hat{f}_\psi$, helper $D_\theta$, and
action-net $E_\phi$ are all updated using surrogate predictions or
replay-buffer mini-batches, so no oracle calls occur inside the
$K_a$, $K_s$, or $K_b$ loops.
The total per-iteration FLOP cost is
$\mathcal{O}(K_s B W_s + K_a B H + K_b B W_s)$,
which is \emph{constant in $T$} --- qualitatively different from all
GP- and BNN-based baselines.
By eliminating the inner-loop acquisition optimizer via the amortised
action-net, acquisition FLOPs reduce from
$\mathcal{O}(N_\text{starts}\cdot N_\text{steps}\cdot S\cdot W)$ to
$\mathcal{O}(K_b\cdot W_e)$, yielding a near $10^2\times$ reduction.
Table~\ref{tab:flops} summarises the comparison; the complexity figure
is in Appendix~\ref{app:additional}.

\begin{table}[t]
\centering
\caption{FLOP and oracle complexity per BO iteration.
  $T$: observations; $W$: BNN weights; $W_s$/$W_e$: surrogate/action-net
  weights; $H$: LSTM dim; $S$: HMC samples; $L$: leapfrog steps.}
\label{tab:flops}
\vskip 0.05in
\resizebox{\columnwidth}{!}{%
\begin{tabular}{lcccc}
\toprule
Method & Surrogate Upd. & Acq.\ Opt. & $f$/iter & Scaling \\
\midrule
GP         & $\mathcal{O}(T^3)$   & $N\mathcal{O}(T^2)$  & $B$ & Cubic \\
HMC-BNN    & $\mathcal{O}(SLTW)$  & $N\mathcal{O}(SW)$   & $B$ & Hi.\ Lin. \\
DKL        & $\mathcal{O}(TW\!+\!T^3)$ & $N\mathcal{O}(T^2)$ & $B$ & Cubic \\
LLA        & $\mathcal{O}(TW\!+\!W^3)$ & $N\mathcal{O}(W^2)$ & $B$ & Linear \\
\textbf{VBO-MI} & $\mathcal{O}(K_sBW_s\!+\!K_aBH)$ & $\mathcal{O}(K_bW_e)$ & $\mathbf{B}$ & \textbf{Const.} \\
\bottomrule
\end{tabular}}
\end{table}

\section{Numerical Evaluation}

We present numerical evaluations of VBO-MI and compare against:
LLA~\cite{pmlr-v130-immer21a,li2024a},
HMC~\cite{pmlr-v139-izmailov21a,li2024a},
SGHMC~\cite{pmlr-v32-cheni14,li2024a},
DKL~\cite{pmlr-v51-wilson16,li2024a},
IBNN~\cite{lee2018deep,li2024a},
GP~\cite{1snoek,li2024a}, and
Focal Sparse GP~\cite{NEURIPS2024_da30215e}.

\paragraph{Implementation.}
VBO-MI uses batch size $B=30$/$60$ for real-world/synthetic tasks,
learning rate $2\!\times\!10^{-3}$ (Adam) for all networks,
$K_a=5$, $K_b=10$, $K_s=10$, and $W=30$ warm-up iterations. Moreover, we set $\beta=10$ in Ackley, and lunar lander, and $\beta=20$ in other tasks.
Baselines follow~\cite{li2024a}.
All results average over 5 seeds; shaded regions show $\pm1$ std. Furthermore, x-axis evaluation represents all queries of the main function $f$ (including the warm-up step in VBO-MI) in different baselines.
Architecture details are in Appendix~\ref{app:setup}.
\begin{figure*}[t]
\centering
\begin{subfigure}{0.48\textwidth}
    \centering
    \includegraphics[width=\linewidth]{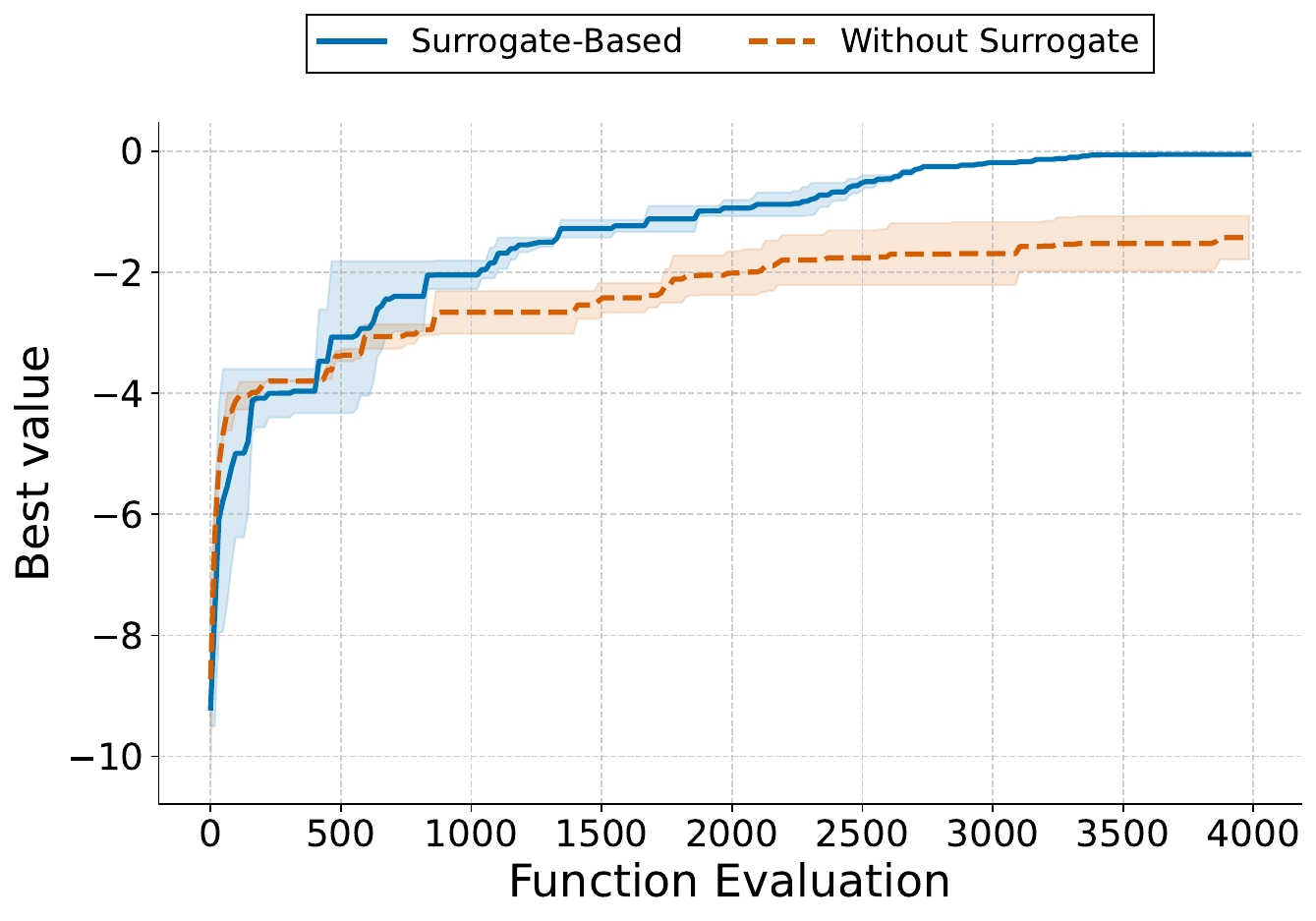}
    \caption{Ackley ($d=10$)}
    \label{fig:ab1}
\end{subfigure}
\hfill
\begin{subfigure}{0.48\textwidth}
    \centering
    \includegraphics[width=\linewidth]{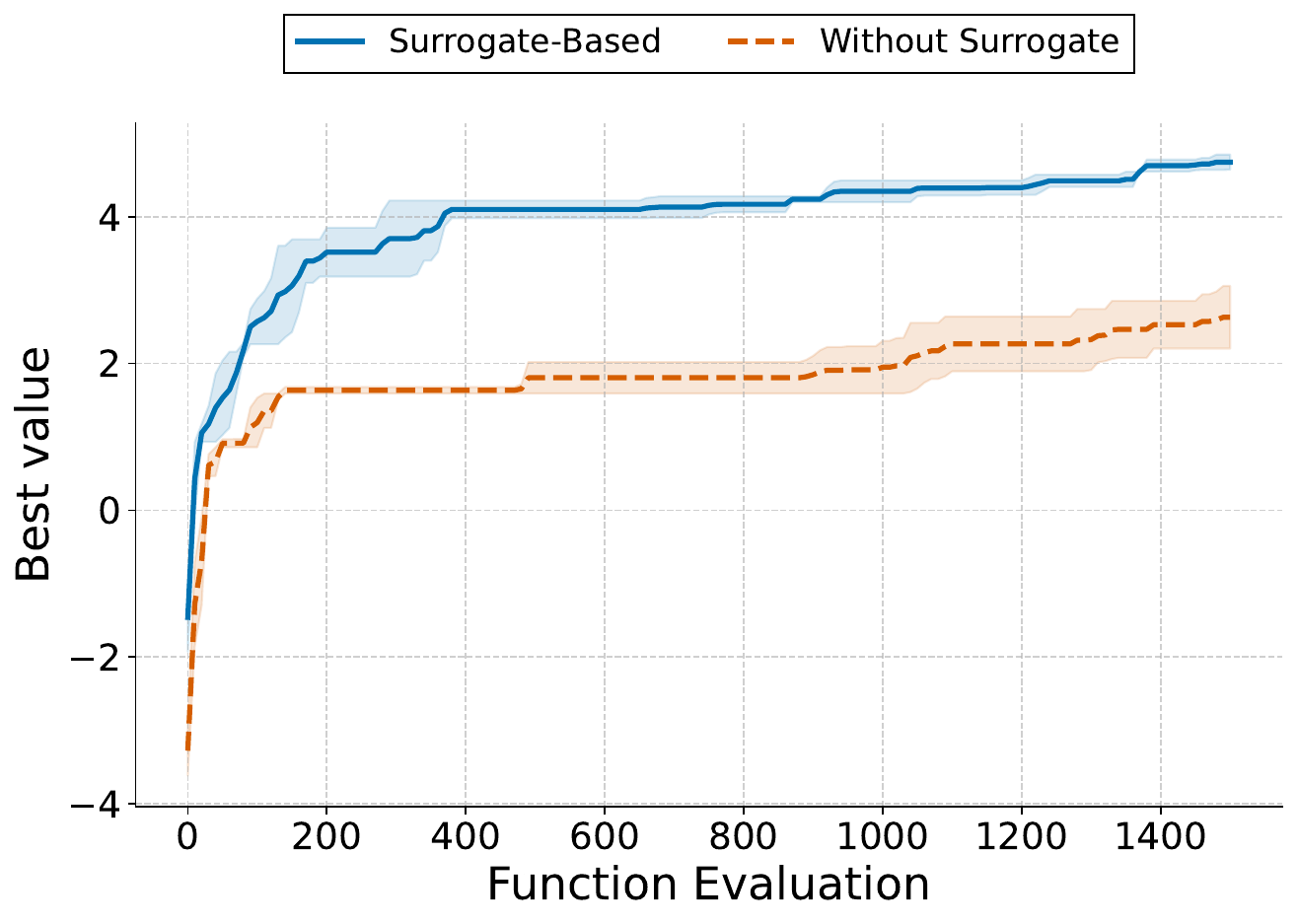}
    \caption{Rover ($d=60$)}
    \label{fig:ab2}
\end{subfigure}
\caption{Ablation~I: surrogate vs.\ direct queries.
         Mean $\pm$ 1 std, 3 trials.}
\label{fig:ablation-surrogate}
\end{figure*}

\begin{figure*}[t]
\centering
\begin{subfigure}{0.48\textwidth}
    \centering
    \includegraphics[width=\linewidth]{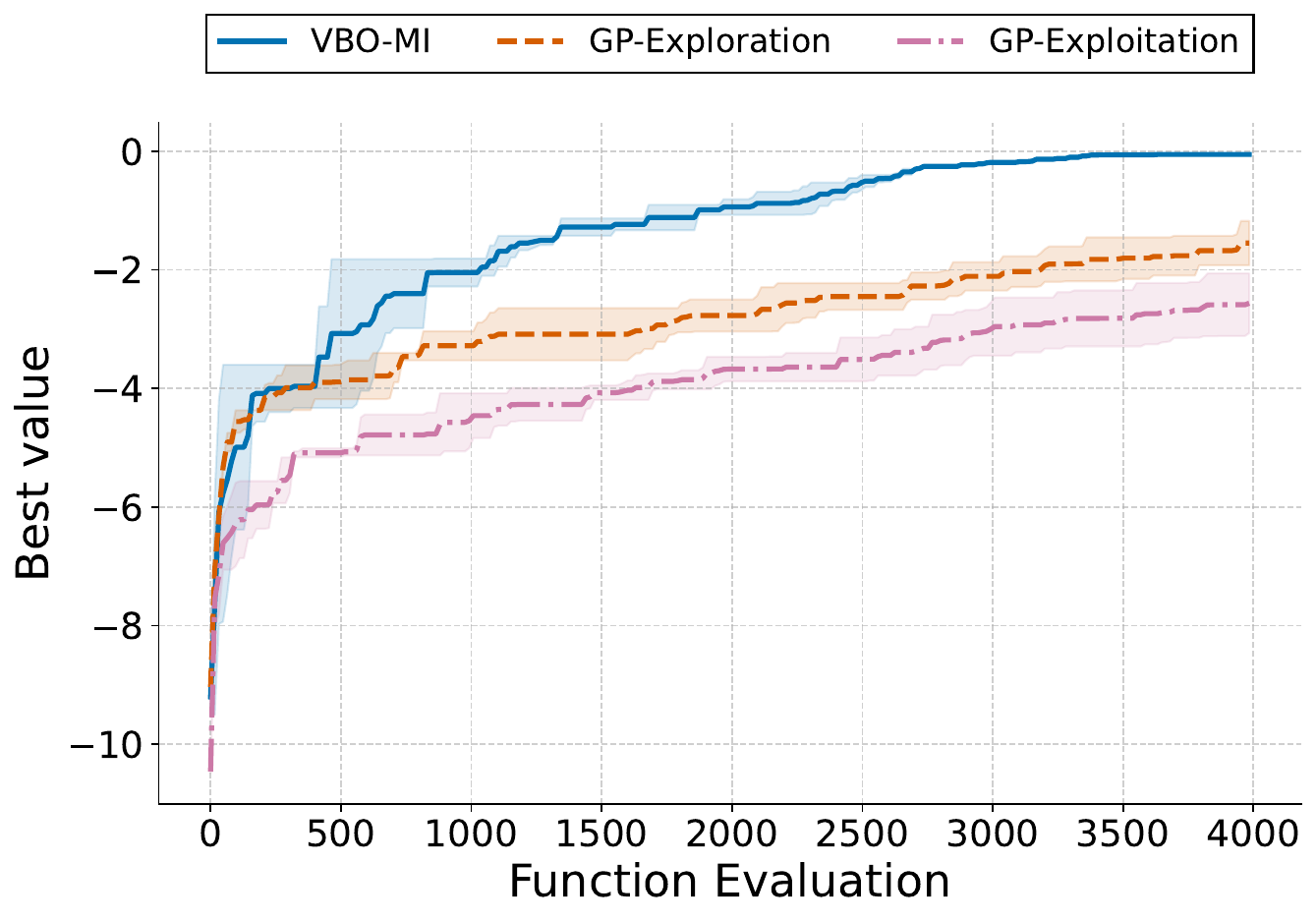}
    \caption{Ackley ($d=10$)}
    \label{fig:ab3}
\end{subfigure}
\hfill
\begin{subfigure}{0.48\textwidth}
    \centering
    \includegraphics[width=\linewidth]{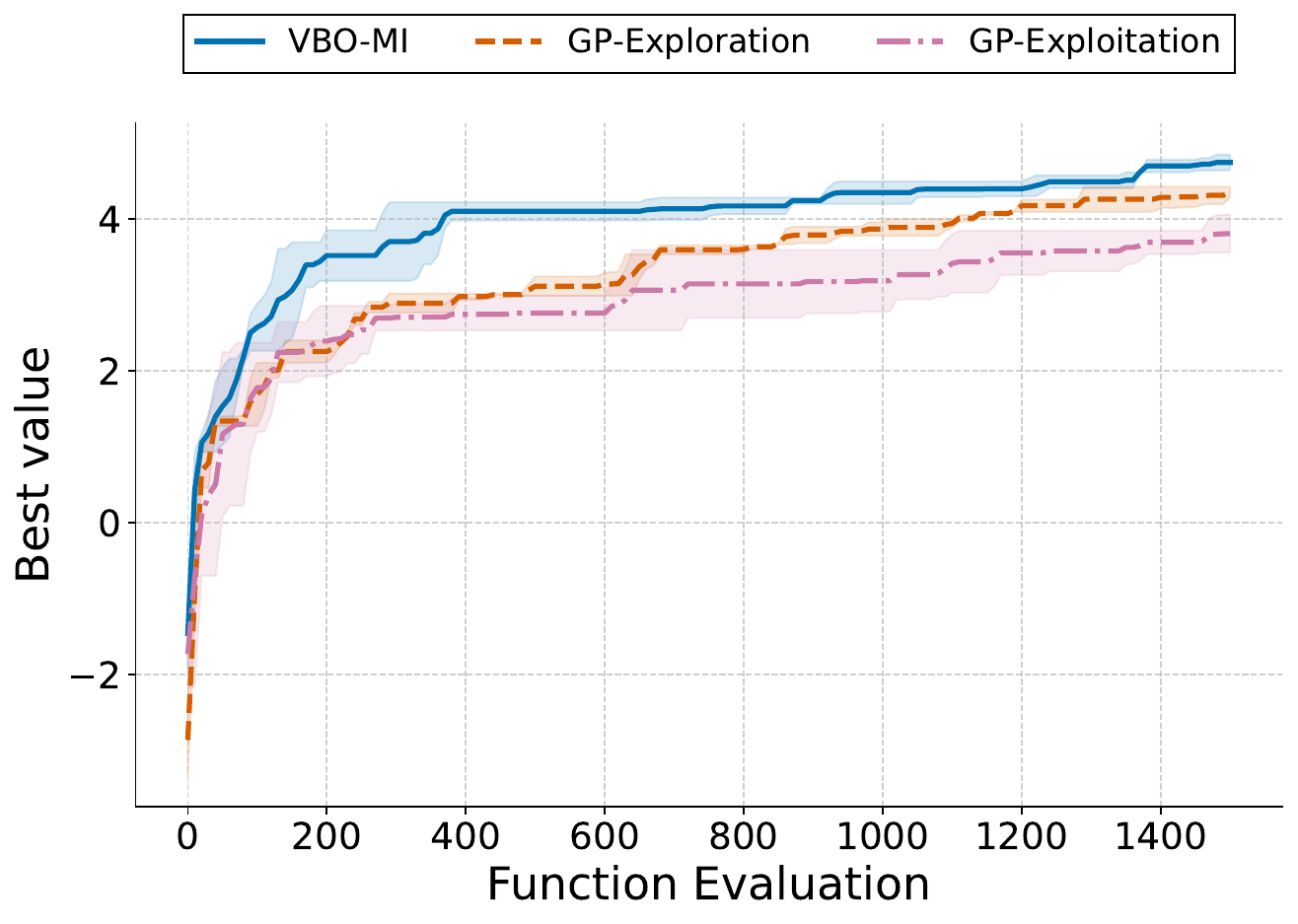}
    \caption{Rover ($d=60$)}
    \label{fig:ab4}
\end{subfigure}
\caption{Ablation~II: acquisition term contribution.
         Full VBO-MI vs.\ GP-exploration and GP-exploitation hybrids.
         Mean $\pm$ 1 std, 3 trials.}
\label{fig:ablation-acquisition}
\end{figure*}

\subsection{Synthetic Functions}

We evaluate on three benchmark functions: Ackley, Levy, and
Griewank~\cite{li2024a,NEURIPS2019_6c990b7a} --- all highly
non-convex with many local optima and global minimum at zero.
We optimize $-f$ so the global maximum is $0$.

\textbf{Ackley ($d=10$, Fig.~\ref{fig:ackley}).}
Ackley is deceptive due to its near-flat outer region and sharp global
basin, making exploration critical.
VBO-MI achieves competitive performance, reaching higher final average
rewards than BNN baselines, and surpasses Focal Sparse GP as iterations
progress.

\textbf{Levy ($d=20$, Fig.~\ref{fig:levy}).}
Levy's difficulty scales with dimension.
VBO-MI demonstrates faster initial convergence and higher final reward
compared to all baselines, including GP and IBNN.

\textbf{Griewank ($d=20$, Fig.~\ref{fig:griewank}).}
Unlike Ackley and Levy, Griewank is \emph{non-separable} due to its
product term, making stationary kernels less effective.
VBO-MI achieves the highest final reward, which we attribute to its
non-parametric variational exploration signal.


\subsection{Real-World Tasks}

We evaluate on three tasks of increasing
dimensionality~\cite{li2024a,NEURIPS2022_12143893}.

\paragraph{Lunar Lander ($d=12$, Fig.~\ref{fig:lunar}).}
The goal is to optimise a controller for OpenAI Gym
\texttt{LunarLander-v2}, maximising average return over 50 random
environments.
VBO-MI achieves the highest final reward with a clear margin over BNN
baselines.

\paragraph{Pest Control ($d=25$, categorical, Fig.~\ref{fig:pest}).}
25 categorical variables, 5 values each~\cite{NEURIPS2019_6c990b7a}.
We apply one-hot encoding~\cite{GARRIDOMERCHAN202020}.
VBO-MI achieves superior performance, with the advantage growing with
more iterations.

\paragraph{Rover Trajectory ($d=60$, Fig.~\ref{fig:rover}).}
30 waypoints in a 2D environment (60 total
dimensions)~\cite{NEURIPS2019_6c990b7a}.
VBO-MI achieves comparable or superior final reward while requiring
significantly fewer oracle evaluations per iteration.



We isolate the contribution of (I) the surrogate network $\hat{f}_\psi$
and (II) each term of the acquisition function $L_{E_\phi}$, evaluated on
Ackley ($d=10$) and Rover ($d=60$) with all other hyperparameters fixed.

\paragraph{Ablation~I: Surrogate network (Fig.~\ref{fig:ablation-surrogate}).}
We compare surrogate-assisted VBO-MI against a variant that queries $f$
directly inside the $K_a$ and $K_b$ loops.
On Ackley the surrogate already provides a higher final reward in early
iterations.
On Rover the advantage is substantially more pronounced: the no-surrogate
variant plateaus at a noticeably lower reward.
With default hyperparameters ($K_a=5$, $K_b=10$), the surrogate reduces
oracle queries from $(K_a+K_b+1)\times B$ to $B$ per iteration --- a
$16\times$ reduction --- at no performance cost.

\paragraph{Ablation~II: Acquisition terms (Fig.~\ref{fig:ablation-acquisition}).}
We replace the exploration term $\sqrt{\beta}\,I_\theta$ with GP posterior
variance $\sqrt{\beta}\,\sigma^2_{\mathrm{GP}}$ (first hybrid), or replace
the exploitation term $\hat{f}_\psi$ with GP posterior mean
$\mu_{\mathrm{GP}}$ (second hybrid).
Full VBO-MI outperforms both hybrids on both tasks.
Replacing the exploitation term produces a larger degradation, particularly
on Rover, confirming that $\hat{f}_\psi$ plays a more critical role than
the exploration signal alone.
Both terms are necessary; neither can be substituted without measurable cost.
We discuss other ablation studies in the appendix on the $beta$ and batch size parameters in the appendix~\ref{ablationf}.

\section{Conclusion}

We presented VBO-MI, a fully gradient-based Bayesian optimization
framework that eliminates GP priors and specific variational family
assumptions.
By combining a variational MI estimator for exploration with a
data-driven surrogate for exploitation, all three networks in VBO-MI
are trained via backpropagation while treating $f$ as a strict black
box.
The surrogate reduces oracle queries to one batch per iteration
(a $16\times$ reduction), and the amortised action-net replaces the
inner-loop acquisition optimizer (a $10^2\times$ FLOP reduction).
We established consistency guarantees and demonstrated competitive or
superior performance on diverse benchmarks spanning $d=10$ to $d=60$.
Future work includes extending VBO-MI to multi-objective settings and
investigating tighter MI estimators for sparse reward landscapes.

\section*{Impact Statement}
This work proposes a general-purpose framework for Bayesian optimization
based on variational mutual information estimation and a neural
actor-critic structure.
As a methodological contribution, it is intended to improve optimization
performance across a wide range of tasks without targeting any specific
application domain.
Bayesian optimization techniques are used in engineering design,
scientific experimentation, and machine learning model tuning, with
potential societal consequences that we do not feel must be specifically
highlighted here.

\bibliographystyle{icml2026}
\bibliography{main}

\appendix

\end{document}